\documentclass[10pt,twocolumn,letterpaper]{article}

\usepackage{cvpr}              
\usepackage[accsupp]{axessibility} 

\usepackage[dvipsnames]{xcolor}

\usepackage{times}
\usepackage{epsfig}
\usepackage{graphicx}
\usepackage{amsmath}
\usepackage{amssymb}
\usepackage{enumitem}
\usepackage{lipsum}
\usepackage{algorithm} 
\usepackage{algorithmic}  
\usepackage[algo2e]{algorithm2e} 
\usepackage[para,online,flushleft]{threeparttable}
\usepackage{makecell}
\usepackage{booktabs}
\usepackage{multirow}

\definecolor{cvprblue}{rgb}{0.21,0.49,0.74}
\usepackage[pagebackref,breaklinks,colorlinks,citecolor=cvprblue]{hyperref}

\newcommand{\M}{SNED}

\def\paperID{10939} 
\def\confName{CVPR}
\def\confYear{2024}

\title{SNED: Superposition Network Architecture Search for Efficient Video Diffusion Model
}

\author{Zhengang Li\textsuperscript{1,2},
Yan Kang\textsuperscript{2}, Yuchen Liu\textsuperscript{2},  Difan Liu\textsuperscript{2}, Tobias Hinz\textsuperscript{3}, Feng Liu\textsuperscript{2}, Yanzhi Wang\textsuperscript{1}\\
\textsuperscript{1}Northeastern University, \textsuperscript{2}Adobe Research, \textsuperscript{3}Adobe\\ 
{\tt\small \textsuperscript{\rm 1}\{li.zhen, yanz.wang\}@northeastern.edu}, 
{\tt\small \textsuperscript{\rm 2,3}\{yankang, yuliu, diliu, thinz, fengl\}@adobe.com}
}

\begin{document}


\maketitle


\begin{abstract}

While AI-generated content has garnered significant attention, achieving photo-realistic video synthesis remains a formidable challenge. Despite the promising advances in diffusion models for video generation quality, the complex model architecture and substantial computational demands for both training and inference create a significant gap between these models and real-world applications. This paper presents~\M, a \textbf{s}uperposition \textbf{n}etwork architecture search method for \textbf{e}fficient video \textbf{d}iffusion model. Our method employs a supernet training paradigm that targets various model cost and resolution options using a weight-sharing method. Moreover, we propose the supernet training sampling warm-up for fast training optimization. To showcase the flexibility of our method, we conduct experiments involving both pixel-space and latent-space video diffusion models. The results demonstrate that our framework consistently produces comparable results across different model options with high efficiency. According to the experiment for the pixel-space video diffusion model, we can achieve consistent video generation results simultaneously across 64$\times$64 to 256$\times$256 resolutions with a large range of model sizes from 640M to 1.6B number of parameters for pixel-space video diffusion models.

\end{abstract}    
\section{Introduction}
\label{sec:intro}
Generative modeling for video synthesis has made tremendous progress based on approaches, including GANs~\cite{yu2022generating, skorokhodov2022stylegan, tian2021good, tulyakov2018mocogan, saito2020train, saito2017temporal, vondrick2016generating}, autoregressive models~\cite{yan2021videogpt, ge2022long}, VAEs~\cite{walker2021predicting, he2018probabilistic}, and normalizing flows~\cite{kumar2019videoflow}. Among them, GANs have demonstrated remarkable success by extending the image-based generation to video generation with dedicated temporal designs. However, GANs encounter challenges such as mode collapse and training instability, making it difficult to scale them up for handling complex and diverse video distributions. 

To overcome this challenge, diffusion models~\cite{ho2020denoising,song2020denoising,dhariwal2021diffusion} have been studied, which establish a weighted variational bound for optimization by connecting Langevin dynamics~\cite{sohl2015deep} and denoising score matching~\cite{song2020score}. Following this, approaches such as VDM~\cite{he2022latent}, MCVD~\cite{voleti2022mcvd}, Imagen Video~\cite{ho2022imagen}, and LVDM~\cite{he2022latent} extended diffusion models to video generation, surpassing GANs in both sample quality and distribution coverage due to their stable training and scalability~\cite{dhariwal2021diffusion}. However, this success comes hand in hand with significant challenges posed by enormous model sizes and computational demands associated with diffusion models. These challenges manifest themselves in both the inference and training aspects. 




Sampling from diffusion models is expensive as their immense number of parameters, heavy reliance on attention mechanisms, and need for several model evaluations result in substantial memory consumption. Even with advanced GPUs, tackling high-resolution video generation becomes a formidable burden due to this memory constraint. Some research efforts, such as Imagen Video~\cite{ho2022imagen}, introduce a model chain to enhance video generation quality gradually. However, this approach further escalates the total model parameters and memory consumption. Beyond this, the extensive computational load leads to a significantly longer inference latency, amplifying the deployment cost and user waiting time. These barriers have presented substantial impediments to the commercialization of diffusion models, particularly in the context of video diffusion models. 

Furthermore, when it comes to the training of diffusion models, challenges emerge from three key facets. Firstly, due to the substantial model parameter count and computational overhead, training costs soar, often requiring an entire month or even longer to train large-scale diffusion models on large datasets from scratch. This protracted training duration poses a challenge to the improvement of diffusion models. Moreover, given that diffusion models are still relatively nascent, our prior knowledge regarding their structural design remains limited. Consequently, model design heavily relies on trial and error, incurring additional expenses in terms of both time and resources. Lastly, because the objectives of these models vary widely, which has different model size constraints and different target video generation resolutions, model architectures often need to be tailored differently to suit each specific goal. Training these diverse models with distinct structures for varying objectives introduces additional overhead that can be burdensome and difficult to manage.

In the face of these challenges, it becomes imperative to explore strategies that mitigate the computational burden and streamline the network design process achieving different targets including cost constraints and resolution requirements at the same time. This is crucial not only for enhancing the efficiency of diffusion models but also for facilitating their broader applicability across various real-world scenarios.



In this paper, we introduce~\M, a \textbf{s}uperposition \textbf{n}etwork architecture search method for \textbf{e}fficient video \textbf{d}iffusion models, designed to achieve efficient model implementation without compromising high-quality generative performance. We explore the combination of network search with video diffusion model and enable a flexible range of options towards resolution and model cost, saving computation consumption for inference and training.
Specifically, we implement a one-shot neural architecture search solution, enabling dynamic computation cost sampling. This means that once the supernet is trained, it achieves the differentiation of computational costs across various subnets within the supernet. Besides that, we introduce the concept of ``super-position training" into our supernet training process. This breakthrough allows a singular supernet model to effectively manage different resolutions, offering a versatile solution for handling diverse resolution requirements. Consequently, this approach permits the reutilization of super-resolution models in multiple instances, facilitating the training of models with a diverse range of cost and resolution options concurrently.

The contributions of this paper include:
\begin{itemize}
    \item A video diffusion model supernet training paradigm that trains subnets with different model sizes and resolution options through a weight-sharing method.
    \item Increasing the search space in different search dimensions including dynamic channels and fine-grained dynamic blocks.
    \item The supernet training sampling warmup strategy to improve the training performance.
    \item Being compatible with different base architectures such as pixel-space and latent-space video diffusion models. 
    \item According to the experiment for pixel-space video diffusion model, we can achieve consistent video generation results simultaneously across 64$\times$64 to 256$\times$256 resolutions with a large range of model sizes from 640M to 1.6B number of parameters for pixel-space video diffusion models.




    
\end{itemize}








\section{Related Work}
\label{sec:relatedwork}

\subsection{Classic Video Synthesis}

Classic video synthesis endeavors to capture the underlying distribution of real-world videos, allowing the generation of realistic and novel video samples. Previous research primarily leverages deep generative models, including GANs~\cite{yu2022generating, skorokhodov2022stylegan, tian2021good, tulyakov2018mocogan, saito2020train, saito2017temporal, vondrick2016generating}, autoregressive models~\cite{yan2021videogpt, ge2022long}, VAEs~\cite{walker2021predicting, he2018probabilistic}, and normalizing flows~\cite{kumar2019videoflow}. Among these, GAN-based approaches stand out as the most dominant, owing to the remarkable success of GANs in image modeling.

MoCoGAN~\cite{tulyakov2018mocogan} and MoCoGAN-HD~\cite{tian2021good} excel in decomposing latent codes into content and motion subspaces. Notably, MoCoGAN-HD~\cite{tian2021good} utilizes the potent pretrained StyleGAN2 as the content generator, resulting in higher-resolution video generation. StyleGAN-V~\cite{skorokhodov2022stylegan} and DiGAN~\cite{yu2022generating} introduce implicit neural representation to GANs, facilitating the modeling of temporal dynamics continuity. These models build upon StyleGAN3 and employ a hierarchical generator architecture for long-range modeling, enabling the generation of videos with evolving content over time.

Despite the success of GANs, they often face challenges such as mode collapse and training instability. Autoregressive methods have also been explored for video generation. VideoGPT~\cite{yan2021videogpt}, utilizing VQVAE~\cite{van2017neural} and a transformer, autoregressively generates tokens in a discrete latent space. TATS~\cite{ge2022long} enhances the VQVAE~\cite{van2017neural} with a more powerful VQGAN~\cite{esser2021taming} and integrates a frame interpolation transformer for rendering long videos in a hierarchical manner.

\subsection{Diffusion Model}

Besides the classic video synthesis models, diffusion models, a category of likelihood-based generative models, have exhibited notable advancements in image and video synthesis tasks, surpassing GANs in both sample quality and distribution coverage due to their stable training and scalability~\cite{dhariwal2021diffusion}. Noteworthy among these models is DDPM~\cite{ho2020denoising}, which establishes a weighted variational bound for optimization by connecting Langevin dynamics~\cite{sohl2015deep} and denoising score matching~\cite{song2020score}. Despite its slow sampling process requiring step-by-step Markov chain progression, DDIM~\cite{song2020denoising} accelerates sampling iteratively in a non-Markovian manner, maintaining the original training process~\cite{nichol2021improved}. ADM~\cite{dhariwal2021diffusion} outperforms GAN-based methods with an intricately designed architecture and classifier guidance.

While diffusion models have excelled in image synthesis, their application to video generation has been limited. VDM~\cite{he2022latent} extends diffusion models to the video domain, introducing modifications such as a spatial-temporal factorized 3D network and image-video joint training. MCVD~\cite{voleti2022mcvd} unifies unconditional video generation and conditional frame prediction through random dropping of conditions during training, akin to the classifier-free guidance approach. Make-A-Video~\cite{singer2022make} and Imagen Video~\cite{ho2022imagen} leverage diffusion models for large-scale video synthesis conditioned on text prompts, which conduct diffusion and denoising processes in pixel space. Besides that, LVDM~\cite{he2022latent} extends the video generation work to the latent space and explores how hierarchical architectures and natural extensions of conditional noise augmentation enable the sampling of long videos. However, the efficiency model optimization of the video diffusion model is still waiting for exploration. In this paper, we further explore the combination of network search with the video diffusion model and enable a flexible range of options toward resolution and model cost.





\subsection{Neural Architecture Search}

\subsubsection{NAS Strategies}

There is a growing trend in designing efficient Deep Neural Networks (DNNs) through Neural Architecture Search (NAS). NAS strategies can be broadly categorized into the following approaches based on their searching strategies. Firstly, Reinforcement Learning (RL) methods, such as~\cite{zoph2016neural, zhong2018practical, zoph2018learning}, utilize recurrent neural networks as predictors to validate the accuracy of child networks over a proxy dataset. Secondly, Evolution methods, exemplified by works~\cite{real2019regularized, real2019aging}, employ a pipeline involving parent initialization, population updating, and the generation and elimination of offspring to discover desired networks. Thirdly, One-Shot NAS, as demonstrated in studies such as~\cite{bender2018understanding, you2020greedynas, guo2020single}, involves training a large one-shot model containing all operations and shares the weight parameters among all candidate models.

Weight-sharing NAS, inspired by the above methodologies, has gained popularity due to its training efficiency~\cite{yu2020bignas, wang2021attentivenas, sahni2021compofa}. In this approach, an over-parameterized supernet is trained with weights shared across all sub-networks in the search space, significantly reducing computational costs during the search.

Although most of the mentioned works primarily focus on traditional Convolutional Neural Network (CNN) architectures, recent studies have extended the scope to include the search for efficient Vision Transformer (ViT) architecture. Examples include Autoformer~\cite{chen2021autoformer}, which entangles the model weights of different ViT blocks in the same layer during supernet training with an efficient weight-sharing strategy. This approach reduces both the training model storage consumption and the overall training time.

\subsubsection{Generation Model NAS}

While generation models have achieved significant success in designing neural architectures, their implementation often demands substantial time, effort, and expert knowledge. For instance,~\cite{karras2019style} devised intricate generators and discriminator backbones to efficiently generate high-resolution images. Recognizing the need to alleviate the burden of network engineering, researchers have explored efficient automatic architecture search techniques for GANs.

In 2019, AutoGAN~\cite{gong2019autogan} introduced an architecture search scheme for GANs utilizing NAS algorithms. It defined a search space to capture deformations in GAN architecture and employed an RNN controller to guide search operations. Later, AutoGAN-Distiller (AGD)~\cite{fu2020autogan} is developed by applying AutoML to GAN compression. AGD performs end-to-end search for efficient generators based on the original GAN model via knowledge distillation. In 2021, alphaGAN~\cite{tian2021alphagan} is introduced, which is a fully differentiable search framework solving bi-level minimax optimization problems. Later, StyleGAN2~\cite{karras2020analyzing} expanded the search space by integrating backbone characteristics. 

While the majority of studies have concentrated on GAN-based generation models, the research realm of video diffusion model NAS remains largely unexplored. Given the substantial computation demands of video diffusion models, there is a critical need to delve into more efficient video diffusion model architecture designs.

\section{Methodology}
\label{sec:methodology}

\subsection{Overview of~\M}

\begin{figure*}[h]
    \centering
    \includegraphics[width=0.95 \textwidth]{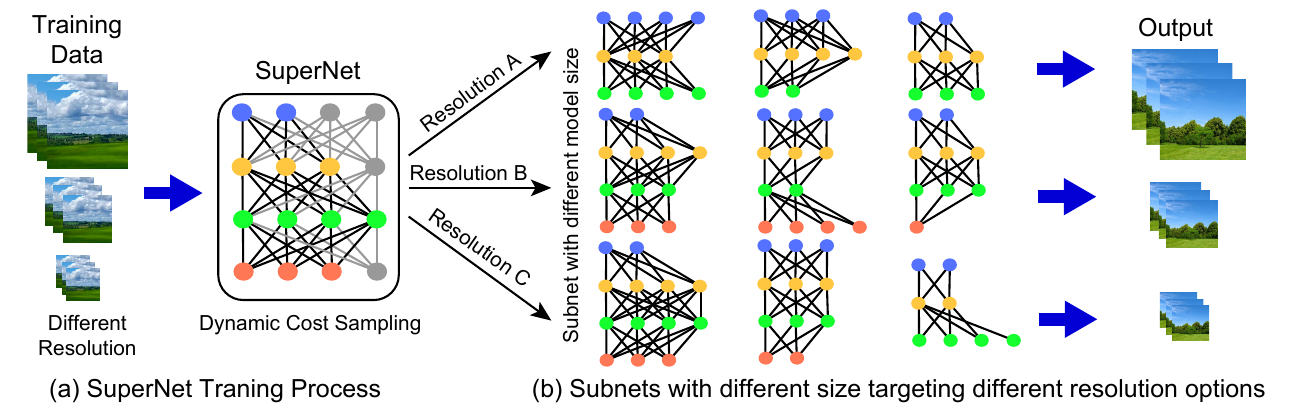}
    \caption{Overview of~\M~framework. (a) We train a supernet with network dynamic cost sampling and multiple input resolution options. In each iteration, a subnet of the supernet is sampled for the training, and other parts (grey) is frozen. (b) After the training, we obtain subnets with different model costs for each resolution option.} 
    \label{fig:overview}
\end{figure*}

\begin{figure}[h]
    \centering
    \includegraphics[width=0.45 \textwidth]{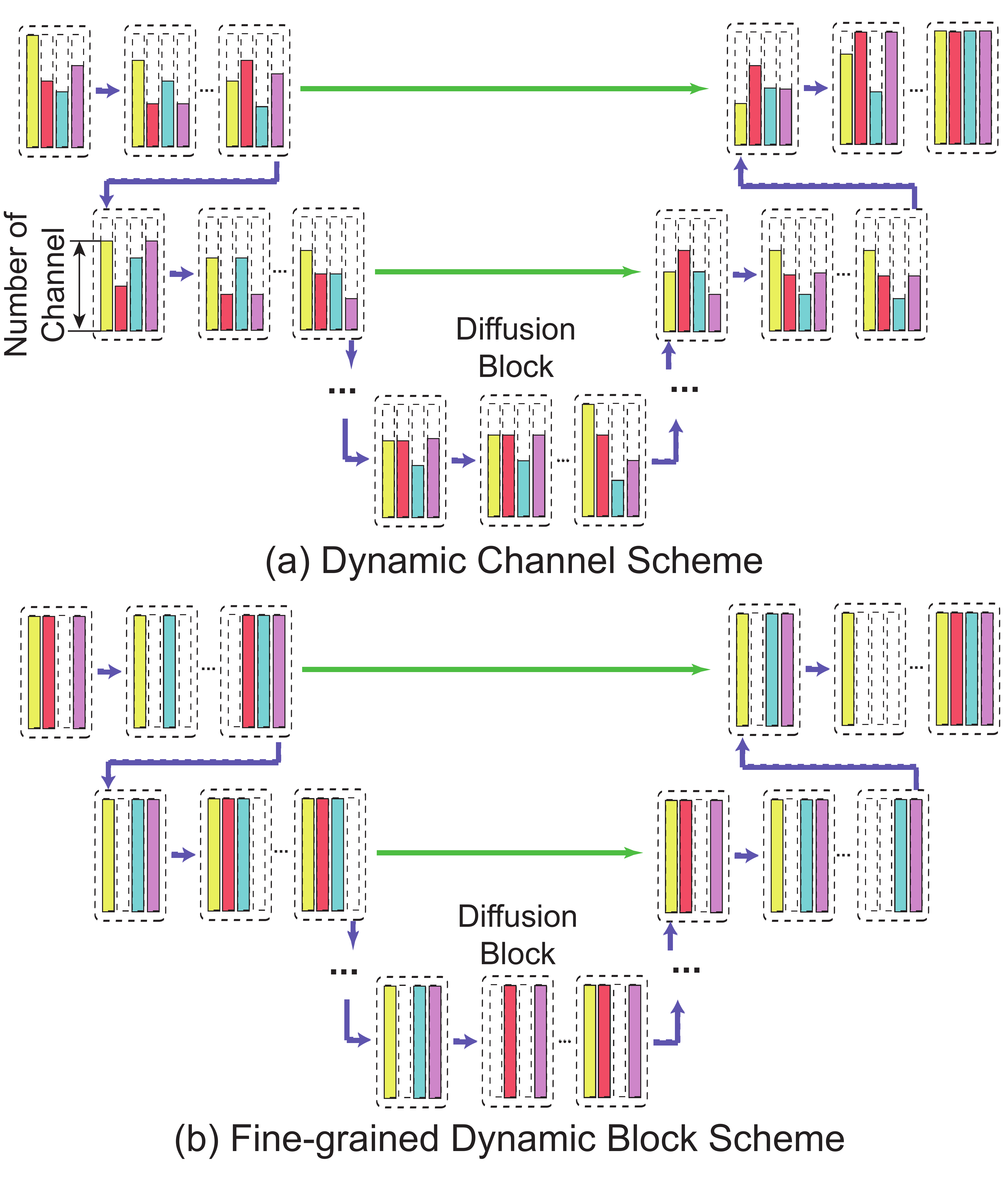}
    \caption{Dynamic cost scheme for~\M~framework.}
    \label{fig:scheme}
\end{figure}

In this paper, we present a framework termed the ``SNED: \textbf{S}uperposition \textbf{N}etwork Architecture Search for \textbf{E}fficient Video \textbf{D}iffusion Models", designed to effectively search for and optimize video diffusion models across multiple dimensions. Our framework introduces two key advancements that address critical challenges in video generation.

The overview framework of~\M~is shown in Fig.~\ref{fig:overview}.
Firstly, we implement a one-shot Neural Architecture Search (NAS) solution, enabling dynamic computation cost sampling. This means that once the supernet is trained, it achieves the differentiation of computational costs across various subnets within the supernet. This feature empowers users to select the appropriate subnetwork based on specific model sizes and computational cost constraints, enhancing flexibility and adaptability. Secondly, we introduce the concept of "super-position NAS training" into our supernet training. This breakthrough allows a singular supernet model to effectively manage different resolutions, offering a versatile solution for handling diverse resolution requirements. Consequently, this approach permits the reutilization of super-resolution models in multiple instances, considerably mitigating the memory overhead within the video diffusion model framework.
By leveraging these advancements, our framework not only streamlines the intricate process of video diffusion model optimization but also substantially reduces memory consumption, paving the way for more efficient and resource-conscious video generation processes.

\subsection{Dynamic Cost Training in~\M}

In each iteration, we randomly select a sampled subnet architecture from the search space and obtain its weights from the supernet. We then compute the losses of the subnet and update the corresponding weights with the remaining supernet weights frozen. The architecture search space $P$ is encoded in a supernet denoted as $\mathcal{S} (P, W_P)$, where $W_P$ is the weight of the supernet that is shared across all the candidate architectures. Algorithm~\ref{alg:supernet_train} illustrates the training procedure of our supernet. 
Here, our dynamic cost search space includes the dynamic channel space and the fine-grained dynamic block space. The schemes of these two search spaces are shown in Fig.~\ref{fig:scheme}. Here, different color bars denote the different components inside a diffusion model, which include ResBlock, temporal self-attention, temporal cross-attention, and spatial attention. The length of the color bars denotes the number of channels of the subnets sampled during training. 

\begin{algorithm}[tb]
\begin{algorithmic}
   \caption{Superposition Supernet Training.} 
   \label{alg:supernet_train}
   \STATE {\bfseries Input:} Training iteration $N$, search space $\mathcal{P}$, supernet $\mathcal{S}$, loss function $L$, train dataset $D_{train}$, initial supernet weights $\mathcal{W}$, candidate weights $\mathcal{W}_p$, Output resolution $R$.
   \FOR{$i$ in $N$ iterations} 
   \FOR{data, labels in $D_{train}$ } 
   \STATE{Randomly sample one subnet architecture and resolution $R$ from search space $\mathcal{P}$. }
   \STATE{Obtain the corresponding weights $\mathcal{W}_p$ from supernet $\mathcal{W}$}
   \STATE{Compute the gradients based on $L$}
  \STATE{ Update the corresponding part of $\mathcal{W}_p$ in $\mathcal{W}$ while freezing the rest of the supernet $\mathcal{S}$}
   \ENDFOR
   \ENDFOR 
   \STATE{\bfseries Output $\mathcal{S}$}
\end{algorithmic}
\end{algorithm}


\noindent\textbf{Dynamic Channel Search Space:}
As the different numbers of channels have different acceleration performances for the hardware implementation, the~\M~search space includes replacing the original number of channels with different percentage ratios, including 100\% (full number of channels), 90\%, 80\%, 70\%, 60\%, 50\%, and 40\%. Each layer inside the diffusion blocks can be assigned an independent ratio in each iteration of supernet training.


\noindent\textbf{Fine-grained Dynamic Block Search space:}
To expand our search space during the supernet training and investigate the potential of the video diffusion model, we add the fine-grained dynamic block search process inside each diffusion block. The basic supernet diffusion block contains four components: ResBlock (convolutional residual block), temporal self-attention block, cross-attention block, and spatial attention block. Our Algorithm enables the drop of a part of the blocks inside the whole diffusion block in each iteration of supernet training. Specifically, if all the attentions inside the diffusion block are dropped, the corresponding feed-forward layer will also be dropped. 

\subsection{Super-position Training in~\M}

We introduce the super-position training mechanism to address different video resolution targets during the supernet training. Here, super-position refers to the utilization of weight-sharing techniques, allowing different subnets to adjust to various resolution processing needs while keeping most of their weights shared. This approach serves the dual purpose of parameter efficiency and the ability to achieve video diffusion models with different resolutions in a single training session.

During each training iteration, besides the sampling of the subnet, we also randomly sample a video generation resolution and preprocess the training data based on that. To balance the training memory workload of different resolution branches, we constraint the maximum model size for different resolutions to ensure an acceptable memory consumption. 
By leveraging this super-position training method, we are not only optimizing the model's resource allocation but also streamlining the training process itself. This minimizes the computational burden and accelerates the development of video diffusion models tailored to different resolution needs.

\subsection{Supernet Training Sampling Warmup}

To achieve a better and faster NAS training performance, we propose the supernet training sampling warmup strategy. This strategy is deployed at the beginning of the supernet training process, improving the supernet's stability and robustness during training.

We gradually increase our search space for both fine-grained dynamic block and dynamic channel during the training, rather than directly applying a full random subnet sampling among the whole search space at the beginning. Specifically, we will apply 30000 training iterations for sampling warmup. The minimum percentage of channels and fine-grained blocks will be decreased from 100\% to 40\% in a step schedule manner.


\section{Experimental Result}
\label{sec:experiment}

\begin{figure*}[h]
    \centering
    \includegraphics[width=0.82 \textwidth]{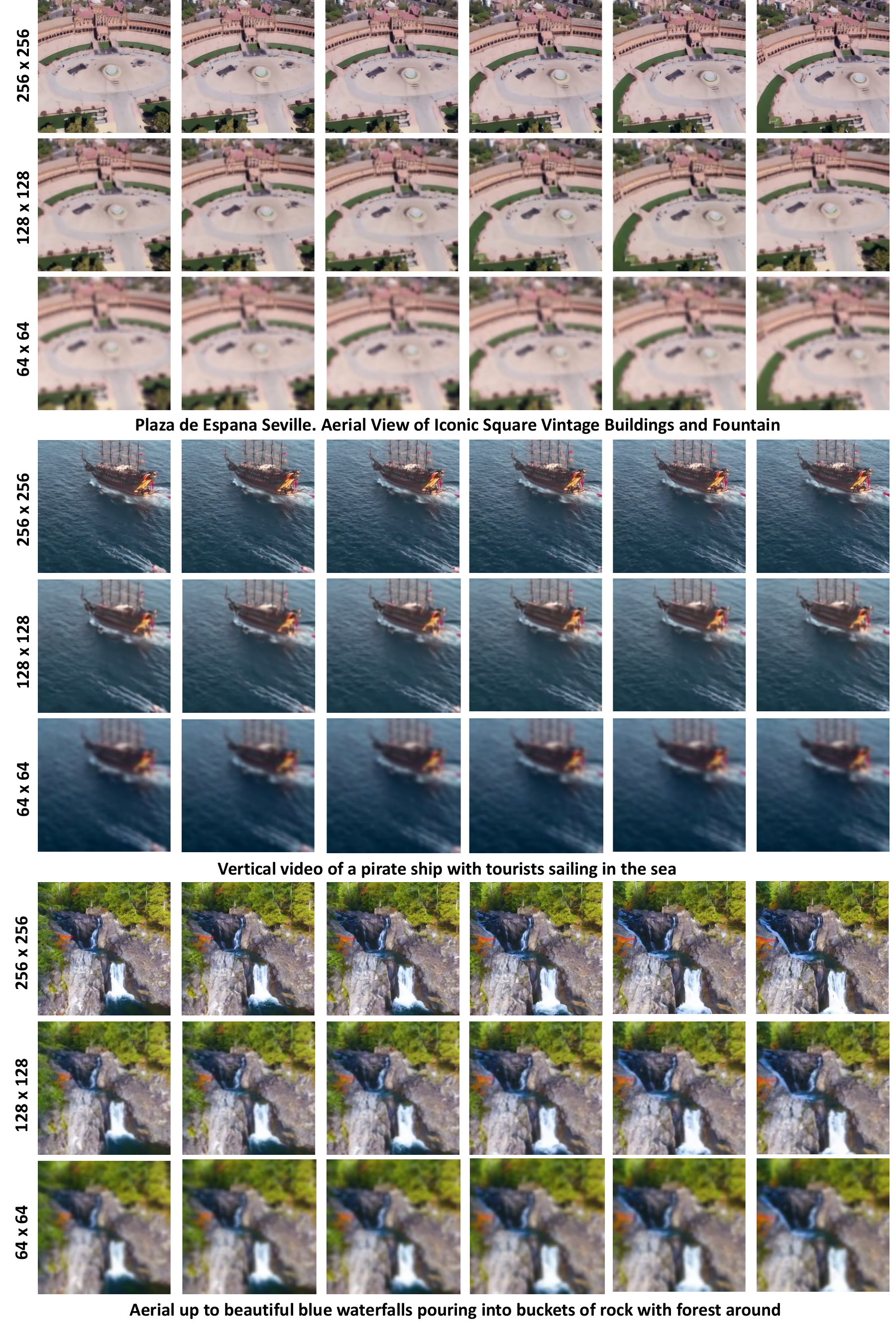}
    \caption{Results of pixel-space video diffusion model for different resolution options. }
    \label{fig:ssr}
\end{figure*}

\subsection{Experimental Setup}


In this section, we present the configuration of our~\M~framework.  Our experiments consist of two primary components: the pixel-space video diffusion model and the latent video diffusion model. To enable the different resolution options under the super-position mechanism, we process the training data into a form suitable for training our cascading pipeline, we spatially resize videos using antialiased bilinear resizing to different resolutions including 64$\times$64, 128$\times$128, and 256$\times$256. To enable the text-to-video conditional training, a frozen text-encoder~\cite{raffel2020exploring} is added at the beginning of the model pipeline.

We train the pixel-space video diffusion model pipeline using an internal dataset comprising 19 million video-text pairs. For the base model and spatial super-resolution (SSR) model inside the pipeline, we use a total batch size of 256 and 64 during training, respectively. Both models undergo 140,000 training iterations, with a fixed learning rate of 0.0001. The training process utilizes 64 A100 GPUs.

For the latent-space video diffusion model, We start with LVDM~\cite{he2022latent} as a baseline and subsequently train it using our algorithm. For a fair comparison, we employ the same publicly available datasets Sky Timelapse. The hyperparameter settings for our experiments align with those of LVDM~\cite{he2022latent} to ensure a fair evaluation.


\subsection{Pixel-Space Video Diffusion Model NAS}

For the pixel-space video diffusion model, our approach is inspired by the model chain proposed by Imagen-video~\cite{ho2022imagen} to realize high-quality video generation. 
The model chain comprises the base model and the spatial super-resolution model (SSR). The base model and SSR are determined by our framework (\M) to meet various computational resource constraints and resolution targets. Our~\M~framework allows for different resolution options in SSR model with weight-sharing subnets. For the supernet architecture of both the base model and SSR model, we apply an imagen-like modified 2D UNet. Each block inside the UNet consists of ResBlock, temporal self-attention and cross-attention, and spatial-attention. 

To attain different resolutions, we recursively deploy our SSR model multiple times instead of integrating multiple SSR models, as demonstrated in Imagen-video~\cite{ho2022imagen}. This approach significantly reduces the total model size.

\subsection{Latent-Space Video Diffusion Model NAS}

Given that Imagen-video does not release its model and dataset, conducting a direct comparison is challenging. To showcase the flexibility and efficiency of our framework, we add additional autoencoder and autodecoder to a latent-space video diffusion model for evaluation. 
The whole model pipeline follows the basic version of LVDM~\cite{he2022latent} \footnote{Since the authors have not released the long-term version training code, we follow and compare it with their released basic latent diffusion model on GitHub. The matrix score is different from that they showed in their paper.}. We first compress video samples to a lower dimensional latent space by the video autoencoder. Then we perform the video generation in the latent space. The encoder and decoder both consist of several layers of 3D convolutions. To ensure that the autoencoder is temporally shift-equivariant, we follow~\cite{he2022latent} to use repeat padding in all three-dimensional convolutions. The prediction model applies a 3D U-Net architecture to estimate the noise distribution, which consists of space-only $1\times3\times3$ shape 3D convolution, and spatial attention module.

We start with LVDM~\cite{he2022latent} as a baseline and subsequently train each part of it inside our NAS framework. Similar to the pixel-space diffusion model, we apply the super-position NAS training to the diffusion prediction model. Since the encoder model is only applied during the training stage, we only apply the super-position training on it without the dynamic cost NAS.

\subsection{Evaluation Results}

\subsubsection{Pixel-Space Video Diffusion Model Visualization}

\begin{figure}[h]
    \centering
    \includegraphics[width=0.48 \textwidth]{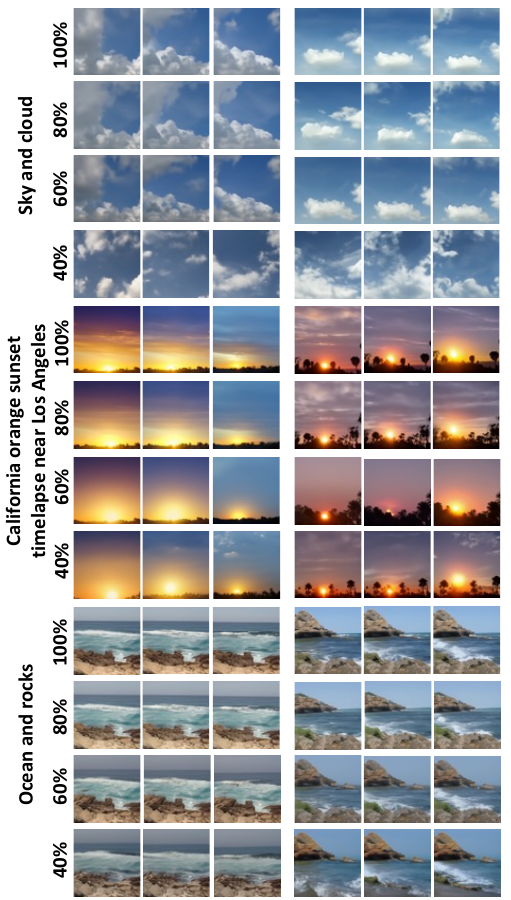}
    \caption{Result of different pixel-space base model subnets with different model sizes. The values of percentage indicate the relative model size compared with the supernet. We show the results of each subnet with two different noise seeds.}
    \label{fig:base}
\end{figure}

In Fig.~\ref{fig:ssr} and Fig.~\ref{fig:base}, we present the results of our pixel-space diffusion model. Fig.~\ref{fig:ssr} shows the generation results from the pixel-space SSR model. We show 6 frames for each of them. Due to the space limitation, we only show the full model size (428M) result of SSR for different resolution options. The corresponding text prompts are listed under each group of video frames. Fig.~\ref{fig:base} illustrates the visualization of the pixel-space base model, transforming the input text (depicted on the left side of the figure) into the corresponding output video. For clarity, we showcase three frames from each video using two different noise seeds. Additionally, for each input text, we display results generated by models of varying sizes—40\% (640M), 60\% (960M), 80\% (1.28B), and 100\% (1.6B) of the parameters compared to the supernet with 1.6B number of parameters. This comprehensive visualization highlights the stability and adaptability of our video generation process achieved through the SNED training strategy.


\begin{figure}[h]
    \centering
    \includegraphics[width=0.48 \textwidth]{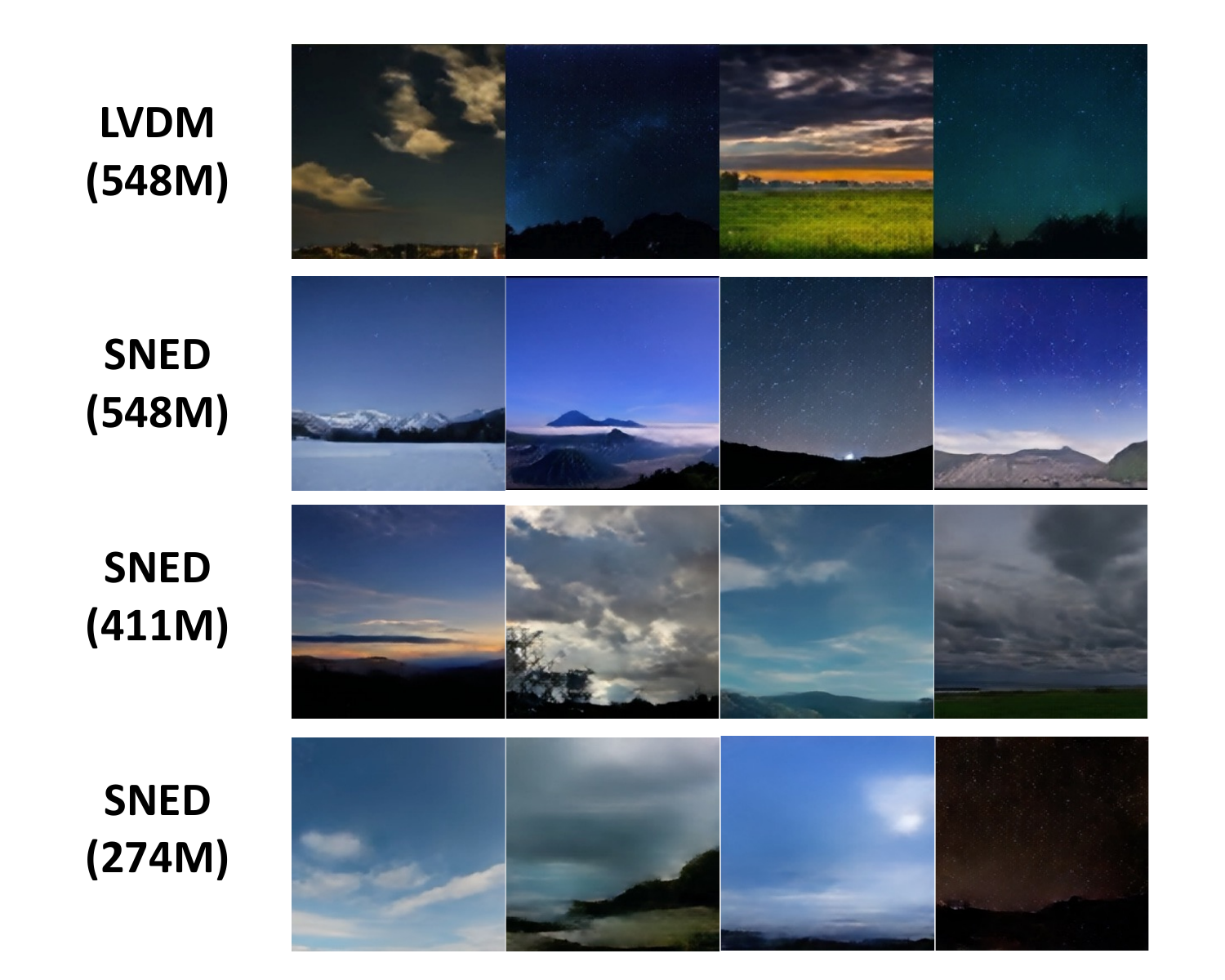}
    \caption{Comparison with LVDM under the resolution of 256$\times$256 on Sky Time-lapse dataset. We present the first frame of each video. Three subnets with different numbers of parameters are included in the comparison.}
    \label{fig:lvdm}
\end{figure}






\subsubsection{Latent-Space Video Diffusion Model Visualization}



The results from the latent-space video diffusion model are depicted in Fig.~\ref{fig:lvdm}. In this comparison, we present the outcomes from three subnets with distinct model sizes (548M, 411M, and 274M) and compare them with the released model from LVDM\cite{he2022latent}. The original output of LVDM\cite{he2022latent} is featured in the first row for reference. All showcased videos in the figure use a consistent resolution of 256$\times$256 and comprise the same number of frames (16) for unconditional short video generation, aligning with the specifications employed in LVDM~\cite{he2022latent}. We show the first frame of each generated video. 

From Fig.~\ref{fig:lvdm}, we can see that, compared with the original LVDM model, all three subnets provide comparable output results for the sky timelapse. Both LVDM and SNED provide generation outputs with high fidelity and diversity, including different skies, clouds, and ground at different times of the day.


\subsection{Model Matrix Evaluation}

For quantitative evaluation, we report the commonly-used FVD~\cite{unterthiner2018towards} and KVD~\cite{unterthiner2018towards} in our experiment. For the pixel-space video diffusion base model, we calculate FVD and KVD scores between 512 real and fake videos with 12 frames, which are presented as $FVD_{12}$ and $KVD_{12}$. All results for the score evaluation are calculated among ten runs to get the average value. The computation is based on the internal dataset comprising 19 million video-text pairs. The latency evaluation is based on one Nvidia A100 GPU. 

\begin{table}[h] 
\centering
\caption{Quantitative comparisons of different subnets for pixel-space video diffusion base model.}
\label{tab:result-base}
\scalebox{0.9}{
\begin{tabular}{c|ccccc}
\toprule
 Model & \#Params (B) & $FVD_{12}\downarrow$  & $KVD_{12}\downarrow$ &  Time (s)  \\ 

\midrule
SNED-B & 1.60 & 544.4 & 25.8 & 24.4\\
SNED-L & 1.28 & 490.5 & 13.0  & 21.2\\
SNED-M & 0.96 & 452.2  & 14.4 & 18.1\\
SNED-S & 0.64 & 472.3  & 16.8 & 16.0   \\
\bottomrule
\end{tabular}}
\label{tab:data}
\end{table}  

As shown in Table~\ref{tab:result-base}, we report the quantitative evaluation for our~\M~models of varying sizes —- small size 40\% (640M), medium size 60\% (960M), large size 80\% (1.28B), and base size 100\% (1.6B) of the parameters compared to the supernet (1.6B), which are indicated as SNED-S, SNED-M, SNED-L, and SNED-B, respectively. From the results, we can see that all of the subnets show a stable score according to both FVD and KVD, which proves our framework's robustness. Small subnets even obtain better FVD and KVD scores compared with the supernet (SNED-B). Among them, SNED-M achieves the best FVD score (452.2), and SNED-L achieves the best KVD score (13.0). Our smallest subnet SNED-S obtains a 472.3 FVD score and a 16.8 KVD score with only 16.0s latency, which achieves 1.53$\times$ of speedup with a better matrix score compared with the supernet model (latency 24.4s).

For the latent-space diffusion model, we compare our matrix score with the baseline LVDM~\cite{he2022latent} and report them in Table~\ref{tab:data}. 
The score computation process follows that used in~\cite{he2022latent}, utilizing 16 frames of generated fake videos for evaluation on the Sky Timelapse dataset. Here we use the released model (548M number of parameters) from LVDM as our supernet architecture, then train it with our dynamic cost schemes. Model size options of 548M, 411M, and 274M are shown in the Table.

\begin{table}[t] 
\centering
\caption{Quantitative comparisons of different subnets under resolution of 256$\times$256.}
\label{tab:result}
\scalebox{0.9}{
\begin{tabular}{c|ccccc}
\toprule
 Model & \#Params (M) & $FVD_{16}\downarrow$  & $KVD_{16}\downarrow$ &  Time (s)  \\ 

\midrule
LVDM & 548 & 295.1 & 20.9 & 86.8\\
SNED & 548 & 298.3 & 20.8  & 86.8\\
SNED & 411 & 348.2  & 23.5 &  74.2\\
SNED & 274 & 472.3  & 28.7 & 66.7   \\
\bottomrule
\end{tabular}}
\label{tab:data}
\end{table}

\section{Conclusion}

This paper introduces~\M, the superposition network architecture search for an efficient video diffusion model. In our training paradigm, we target various model cost and resolution options using a weight-sharing method and incorporate both dynamic channel and fine-grained dynamic block to expand our search space. Additionally, we propose the supernet training sampling warmup to improve the training performance. Our proposed method is compatible with different base architectures such as pixel-space and latent-space video diffusion models. 
According to the experimental results for the pixel-space video diffusion model, we can achieve consistent video generation results simultaneously across 64$\times$64 to 256$\times$256 resolutions with a large model size range from 640M to 1.6B number of parameters. To the best of our knowledge, this is the first NAS framework targeting the video diffusion model. 

\section{Acknowledgement}

This work was completed during Zhengang's internship at Adobe Research and was partially supported by the National Science Foundation under grant number CNS-1909172.

{
    \small
    \bibliographystyle{ieeenat_fullname}
    \bibliography{main}

\begin{thebibliography}{43}
\providecommand{\natexlab}[1]{#1}
\providecommand{\url}[1]{\texttt{#1}}
\expandafter\ifx\csname urlstyle\endcsname\relax
  \providecommand{\doi}[1]{doi: #1}\else
  \providecommand{\doi}{doi: \begingroup \urlstyle{rm}\Url}\fi

\bibitem[Bender et~al.(2018)Bender, Kindermans, Zoph, Vasudevan, and Le]{bender2018understanding}
Gabriel Bender, Pieter-Jan Kindermans, Barret Zoph, Vijay Vasudevan, and Quoc Le.
\newblock Understanding and simplifying one-shot architecture search.
\newblock In \emph{International Conference on Machine Learning}, pages 550--559, 2018.

\bibitem[Chen et~al.(2021)Chen, Peng, Fu, and Ling]{chen2021autoformer}
Minghao Chen, Houwen Peng, Jianlong Fu, and Haibin Ling.
\newblock Autoformer: Searching transformers for visual recognition.
\newblock \emph{arXiv preprint arXiv:2107.00651}, 2021.

\bibitem[Dhariwal and Nichol(2021)]{dhariwal2021diffusion}
Prafulla Dhariwal and Alexander Nichol.
\newblock Diffusion models beat gans on image synthesis.
\newblock \emph{Advances in neural information processing systems}, 34:\penalty0 8780--8794, 2021.

\bibitem[Esser et~al.(2021)Esser, Rombach, and Ommer]{esser2021taming}
Patrick Esser, Robin Rombach, and Bjorn Ommer.
\newblock Taming transformers for high-resolution image synthesis.
\newblock In \emph{Proceedings of the IEEE/CVF conference on computer vision and pattern recognition}, pages 12873--12883, 2021.

\bibitem[Fu et~al.(2020)Fu, Chen, Wang, Li, Lin, and Wang]{fu2020autogan}
Yonggan Fu, Wuyang Chen, Haotao Wang, Haoran Li, Yingyan Lin, and Zhangyang Wang.
\newblock Autogan-distiller: Searching to compress generative adversarial networks.
\newblock \emph{arXiv preprint arXiv:2006.08198}, 2020.

\bibitem[Ge et~al.(2022)Ge, Hayes, Yang, Yin, Pang, Jacobs, Huang, and Parikh]{ge2022long}
Songwei Ge, Thomas Hayes, Harry Yang, Xi Yin, Guan Pang, David Jacobs, Jia-Bin Huang, and Devi Parikh.
\newblock Long video generation with time-agnostic vqgan and time-sensitive transformer.
\newblock In \emph{European Conference on Computer Vision}, pages 102--118. Springer, 2022.

\bibitem[Gong et~al.(2019)Gong, Chang, Jiang, and Wang]{gong2019autogan}
Xinyu Gong, Shiyu Chang, Yifan Jiang, and Zhangyang Wang.
\newblock Autogan: Neural architecture search for generative adversarial networks.
\newblock In \emph{Proceedings of the IEEE/CVF International Conference on Computer Vision}, pages 3224--3234, 2019.

\bibitem[Guo et~al.(2020)Guo, Zhang, Mu, Heng, Liu, Wei, and Sun]{guo2020single}
Zichao Guo, Xiangyu Zhang, Haoyuan Mu, Wen Heng, Zechun Liu, Yichen Wei, and Jian Sun.
\newblock Single path one-shot neural architecture search with uniform sampling.
\newblock In \emph{European Conference on Computer Vision}, pages 544--560. Springer, 2020.

\bibitem[He et~al.(2018)He, Lehrmann, Marino, Mori, and Sigal]{he2018probabilistic}
Jiawei He, Andreas Lehrmann, Joseph Marino, Greg Mori, and Leonid Sigal.
\newblock Probabilistic video generation using holistic attribute control.
\newblock In \emph{Proceedings of the European Conference on Computer Vision (ECCV)}, pages 452--467, 2018.

\bibitem[He et~al.(2022)He, Yang, Zhang, Shan, and Chen]{he2022latent}
Yingqing He, Tianyu Yang, Yong Zhang, Ying Shan, and Qifeng Chen.
\newblock Latent video diffusion models for high-fidelity video generation with arbitrary lengths.
\newblock \emph{arXiv preprint arXiv:2211.13221}, 2022.

\bibitem[Ho et~al.(2020)Ho, Jain, and Abbeel]{ho2020denoising}
Jonathan Ho, Ajay Jain, and Pieter Abbeel.
\newblock Denoising diffusion probabilistic models.
\newblock \emph{Advances in neural information processing systems}, 33:\penalty0 6840--6851, 2020.

\bibitem[Ho et~al.(2022)Ho, Chan, Saharia, Whang, Gao, Gritsenko, Kingma, Poole, Norouzi, Fleet, et~al.]{ho2022imagen}
Jonathan Ho, William Chan, Chitwan Saharia, Jay Whang, Ruiqi Gao, Alexey Gritsenko, Diederik~P Kingma, Ben Poole, Mohammad Norouzi, David~J Fleet, et~al.
\newblock Imagen video: High definition video generation with diffusion models.
\newblock \emph{arXiv preprint arXiv:2210.02303}, 2022.

\bibitem[Karras et~al.(2019)Karras, Laine, and Aila]{karras2019style}
Tero Karras, Samuli Laine, and Timo Aila.
\newblock A style-based generator architecture for generative adversarial networks.
\newblock In \emph{Proceedings of the IEEE/CVF conference on computer vision and pattern recognition}, pages 4401--4410, 2019.

\bibitem[Karras et~al.(2020)Karras, Laine, Aittala, Hellsten, Lehtinen, and Aila]{karras2020analyzing}
Tero Karras, Samuli Laine, Miika Aittala, Janne Hellsten, Jaakko Lehtinen, and Timo Aila.
\newblock Analyzing and improving the image quality of stylegan.
\newblock In \emph{Proceedings of the IEEE/CVF conference on computer vision and pattern recognition}, pages 8110--8119, 2020.

\bibitem[Kumar et~al.(2019)Kumar, Babaeizadeh, Erhan, Finn, Levine, Dinh, and Kingma]{kumar2019videoflow}
Manoj Kumar, Mohammad Babaeizadeh, Dumitru Erhan, Chelsea Finn, Sergey Levine, Laurent Dinh, and Durk Kingma.
\newblock Videoflow: A conditional flow-based model for stochastic video generation.
\newblock \emph{arXiv preprint arXiv:1903.01434}, 2019.

\bibitem[Nichol and Dhariwal(2021)]{nichol2021improved}
Alexander~Quinn Nichol and Prafulla Dhariwal.
\newblock Improved denoising diffusion probabilistic models.
\newblock In \emph{International Conference on Machine Learning}, pages 8162--8171. PMLR, 2021.

\bibitem[Raffel et~al.(2020)Raffel, Shazeer, Roberts, Lee, Narang, Matena, Zhou, Li, and Liu]{raffel2020exploring}
Colin Raffel, Noam Shazeer, Adam Roberts, Katherine Lee, Sharan Narang, Michael Matena, Yanqi Zhou, Wei Li, and Peter~J Liu.
\newblock Exploring the limits of transfer learning with a unified text-to-text transformer.
\newblock \emph{The Journal of Machine Learning Research}, 21\penalty0 (1):\penalty0 5485--5551, 2020.

\bibitem[Real et~al.(2019{\natexlab{a}})Real, Aggarwal, Huang, and Le]{real2019aging}
Esteban Real, Alok Aggarwal, Yanping Huang, and Quoc~V Le.
\newblock Aging evolution for image classifier architecture search.
\newblock In \emph{AAAI Conference on Artificial Intelligence}, 2019{\natexlab{a}}.

\bibitem[Real et~al.(2019{\natexlab{b}})Real, Aggarwal, Huang, and Le]{real2019regularized}
Esteban Real, Alok Aggarwal, Yanping Huang, and Quoc~V Le.
\newblock Regularized evolution for image classifier architecture search.
\newblock In \emph{Proceedings of the aaai conference on artificial intelligence}, pages 4780--4789, 2019{\natexlab{b}}.

\bibitem[Sahni et~al.(2021)Sahni, Varshini, Khare, and Tumanov]{sahni2021compofa}
Manas Sahni, Shreya Varshini, Alind Khare, and Alexey Tumanov.
\newblock Compofa: Compound once-for-all networks for faster multi-platform deployment.
\newblock \emph{arXiv preprint arXiv:2104.12642}, 2021.

\bibitem[Saito et~al.(2017)Saito, Matsumoto, and Saito]{saito2017temporal}
Masaki Saito, Eiichi Matsumoto, and Shunta Saito.
\newblock Temporal generative adversarial nets with singular value clipping.
\newblock In \emph{Proceedings of the IEEE international conference on computer vision}, pages 2830--2839, 2017.

\bibitem[Saito et~al.(2020)Saito, Saito, Koyama, and Kobayashi]{saito2020train}
Masaki Saito, Shunta Saito, Masanori Koyama, and Sosuke Kobayashi.
\newblock Train sparsely, generate densely: Memory-efficient unsupervised training of high-resolution temporal gan.
\newblock \emph{International Journal of Computer Vision}, 128\penalty0 (10-11):\penalty0 2586--2606, 2020.

\bibitem[Singer et~al.(2022)Singer, Polyak, Hayes, Yin, An, Zhang, Hu, Yang, Ashual, Gafni, et~al.]{singer2022make}
Uriel Singer, Adam Polyak, Thomas Hayes, Xi Yin, Jie An, Songyang Zhang, Qiyuan Hu, Harry Yang, Oron Ashual, Oran Gafni, et~al.
\newblock Make-a-video: Text-to-video generation without text-video data.
\newblock \emph{arXiv preprint arXiv:2209.14792}, 2022.

\bibitem[Skorokhodov et~al.(2022)Skorokhodov, Tulyakov, and Elhoseiny]{skorokhodov2022stylegan}
Ivan Skorokhodov, Sergey Tulyakov, and Mohamed Elhoseiny.
\newblock Stylegan-v: A continuous video generator with the price, image quality and perks of stylegan2.
\newblock In \emph{Proceedings of the IEEE/CVF Conference on Computer Vision and Pattern Recognition}, pages 3626--3636, 2022.

\bibitem[Sohl-Dickstein et~al.(2015)Sohl-Dickstein, Weiss, Maheswaranathan, and Ganguli]{sohl2015deep}
Jascha Sohl-Dickstein, Eric Weiss, Niru Maheswaranathan, and Surya Ganguli.
\newblock Deep unsupervised learning using nonequilibrium thermodynamics.
\newblock In \emph{International conference on machine learning}, pages 2256--2265. PMLR, 2015.

\bibitem[Song et~al.(2020{\natexlab{a}})Song, Meng, and Ermon]{song2020denoising}
Jiaming Song, Chenlin Meng, and Stefano Ermon.
\newblock Denoising diffusion implicit models.
\newblock \emph{arXiv preprint arXiv:2010.02502}, 2020{\natexlab{a}}.

\bibitem[Song et~al.(2020{\natexlab{b}})Song, Sohl-Dickstein, Kingma, Kumar, Ermon, and Poole]{song2020score}
Yang Song, Jascha Sohl-Dickstein, Diederik~P Kingma, Abhishek Kumar, Stefano Ermon, and Ben Poole.
\newblock Score-based generative modeling through stochastic differential equations.
\newblock \emph{arXiv preprint arXiv:2011.13456}, 2020{\natexlab{b}}.

\bibitem[Tian et~al.(2021{\natexlab{a}})Tian, Ren, Chai, Olszewski, Peng, Metaxas, and Tulyakov]{tian2021good}
Yu Tian, Jian Ren, Menglei Chai, Kyle Olszewski, Xi Peng, Dimitris~N Metaxas, and Sergey Tulyakov.
\newblock A good image generator is what you need for high-resolution video synthesis.
\newblock \emph{arXiv preprint arXiv:2104.15069}, 2021{\natexlab{a}}.

\bibitem[Tian et~al.(2021{\natexlab{b}})Tian, Shen, Su, Li, and Liu]{tian2021alphagan}
Yuesong Tian, Li Shen, Guinan Su, Zhifeng Li, and Wei Liu.
\newblock Alphagan: Fully differentiable architecture search for generative adversarial networks.
\newblock \emph{IEEE Transactions on Pattern Analysis and Machine Intelligence}, 44\penalty0 (10):\penalty0 6752--6766, 2021{\natexlab{b}}.

\bibitem[Tulyakov et~al.(2018)Tulyakov, Liu, Yang, and Kautz]{tulyakov2018mocogan}
Sergey Tulyakov, Ming-Yu Liu, Xiaodong Yang, and Jan Kautz.
\newblock Mocogan: Decomposing motion and content for video generation.
\newblock In \emph{Proceedings of the IEEE conference on computer vision and pattern recognition}, pages 1526--1535, 2018.

\bibitem[Unterthiner et~al.(2018)Unterthiner, Van~Steenkiste, Kurach, Marinier, Michalski, and Gelly]{unterthiner2018towards}
Thomas Unterthiner, Sjoerd Van~Steenkiste, Karol Kurach, Raphael Marinier, Marcin Michalski, and Sylvain Gelly.
\newblock Towards accurate generative models of video: A new metric \& challenges.
\newblock \emph{arXiv preprint arXiv:1812.01717}, 2018.

\bibitem[Van Den~Oord et~al.(2017)Van Den~Oord, Vinyals, et~al.]{van2017neural}
Aaron Van Den~Oord, Oriol Vinyals, et~al.
\newblock Neural discrete representation learning.
\newblock \emph{Advances in neural information processing systems}, 30, 2017.

\bibitem[Voleti et~al.(2022)Voleti, Jolicoeur-Martineau, and Pal]{voleti2022mcvd}
Vikram Voleti, Alexia Jolicoeur-Martineau, and Chris Pal.
\newblock Mcvd-masked conditional video diffusion for prediction, generation, and interpolation.
\newblock \emph{Advances in Neural Information Processing Systems}, 35:\penalty0 23371--23385, 2022.

\bibitem[Vondrick et~al.(2016)Vondrick, Pirsiavash, and Torralba]{vondrick2016generating}
Carl Vondrick, Hamed Pirsiavash, and Antonio Torralba.
\newblock Generating videos with scene dynamics.
\newblock \emph{Advances in neural information processing systems}, 29, 2016.

\bibitem[Walker et~al.(2021)Walker, Razavi, and Oord]{walker2021predicting}
Jacob Walker, Ali Razavi, and A{\"a}ron van~den Oord.
\newblock Predicting video with vqvae.
\newblock \emph{arXiv preprint arXiv:2103.01950}, 2021.

\bibitem[Wang et~al.(2021)Wang, Li, Gong, and Chandra]{wang2021attentivenas}
Dilin Wang, Meng Li, Chengyue Gong, and Vikas Chandra.
\newblock Attentivenas: Improving neural architecture search via attentive sampling.
\newblock In \emph{Proceedings of the IEEE/CVF Conference on Computer Vision and Pattern Recognition}, pages 6418--6427, 2021.

\bibitem[Yan et~al.(2021)Yan, Zhang, Abbeel, and Srinivas]{yan2021videogpt}
Wilson Yan, Yunzhi Zhang, Pieter Abbeel, and Aravind Srinivas.
\newblock Videogpt: Video generation using vq-vae and transformers.
\newblock \emph{arXiv preprint arXiv:2104.10157}, 2021.

\bibitem[You et~al.(2020)You, Huang, Yang, Wang, Qian, and Zhang]{you2020greedynas}
Shan You, Tao Huang, Mingmin Yang, Fei Wang, Chen Qian, and Changshui Zhang.
\newblock Greedynas: Towards fast one-shot nas with greedy supernet.
\newblock In \emph{Proceedings of the IEEE/CVF Conference on Computer Vision and Pattern Recognition}, pages 1999--2008, 2020.

\bibitem[Yu et~al.(2020)Yu, Jin, Liu, Bender, Kindermans, Tan, Huang, Song, Pang, and Le]{yu2020bignas}
Jiahui Yu, Pengchong Jin, Hanxiao Liu, Gabriel Bender, Pieter-Jan Kindermans, Mingxing Tan, Thomas Huang, Xiaodan Song, Ruoming Pang, and Quoc Le.
\newblock Bignas: Scaling up neural architecture search with big single-stage models.
\newblock In \emph{European Conference on Computer Vision}, pages 702--717. Springer, 2020.

\bibitem[Yu et~al.(2022)Yu, Tack, Mo, Kim, Kim, Ha, and Shin]{yu2022generating}
Sihyun Yu, Jihoon Tack, Sangwoo Mo, Hyunsu Kim, Junho Kim, Jung-Woo Ha, and Jinwoo Shin.
\newblock Generating videos with dynamics-aware implicit generative adversarial networks.
\newblock \emph{arXiv preprint arXiv:2202.10571}, 2022.

\bibitem[Zhong et~al.(2018)Zhong, Yan, Wu, Shao, and Liu]{zhong2018practical}
Zhao Zhong, Junjie Yan, Wei Wu, Jing Shao, and Cheng-Lin Liu.
\newblock Practical block-wise neural network architecture generation.
\newblock In \emph{Proceedings of the IEEE conference on computer vision and pattern recognition}, pages 2423--2432, 2018.

\bibitem[Zoph and Le(2017)]{zoph2016neural}
Barret Zoph and Quoc~V. Le.
\newblock Neural architecture search with reinforcement learning.
\newblock In \emph{International Conference on Learning Representations (ICLR)}, 2017.

\bibitem[Zoph et~al.(2018)Zoph, Vasudevan, Shlens, and Le]{zoph2018learning}
Barret Zoph, Vijay Vasudevan, Jonathon Shlens, and Quoc~V Le.
\newblock Learning transferable architectures for scalable image recognition.
\newblock In \emph{Proceedings of the IEEE conference on computer vision and pattern recognition (CVPR)}, pages 8697--8710, 2018.

\end{thebibliography}
}





\end{document}